\pdfoutput=1

\documentclass[11pt]{article}

\usepackage{EMNLP2023}

\usepackage{times}
\usepackage{latexsym}
\usepackage{graphicx}

\usepackage[T1]{fontenc}

\usepackage[utf8]{inputenc}

\usepackage{microtype}

\usepackage{inconsolata}

\title{MUST\&P-SRL: Multi-lingual and Unified Syllabification in Text and Phonetic Domains for Speech Representation Learning}

\author{Noé Tits \\
  Flowchase \\
  ISIA Lab, Numediart Institute, University of Mons \\
  \texttt{noe.tits@flowchase.app} \\
  \texttt{noe.tits@alumni.umons.ac.be} 
  }

\begin{document}
\maketitle
\begin{abstract}
In this paper, we present a methodology for linguistic feature extraction, focusing particularly on automatically syllabifying words in multiple languages, with a design to be compatible with a forced-alignment tool, the Montreal Forced Aligner (MFA). In both the textual and phonetic domains, our method focuses on the extraction of phonetic transcriptions from text, stress marks, and a unified automatic syllabification (in text and phonetic domains). The system was built with open-source components and resources. Through an ablation study, we demonstrate the efficacy of our approach in automatically syllabifying words from several languages (English, French and Spanish). Additionally, we apply the technique to the transcriptions of the CMU ARCTIC dataset, generating valuable annotations available online\footnote{\url{https://github.com/noetits/MUST_P-SRL}} that are ideal for speech representation learning, speech unit discovery, and disentanglement of speech factors in several speech-related fields.
\end{abstract}

\section{Introduction}

Modern speech technologies have moved towards end-to-end models that constitutes black box systems that do not allow for explainability of the prediction or decisions. This lack of explainability started to raise a lot of concerns in the industry because of the need of identifying causes or reasons for decisions. This lead to the advent of the concept of Explainable AI (XAI) for which the goal is to discover ways to explains why a certain prediction was made by a system.

For this, one avenue is the field of representation learning
which incorporate unsupervised/self-supervised learning, aiming to discover robust and meaningful representations for various tasks and analyze their relationship with expert knowledge (e.g. \cite{visualization-19-tits, tits2021analysis}).
It is well known that in Deep Learning, learning knowledge can be tranferable from one task to other and Self-Supervised Learning is probably the most versatile Transfer Learning technique today.
Transfer Learning~\cite{tan2018survey_deep_transfer_learning} is a widely used technique in Deep Learning for leveraging models trained on related tasks for which there exist abundant datasets towards tasks for which few labels exist.

This principle has been applied successfully for speech technology application~\cite{wang2015transfer} with few available data such as speech recognition for low resource languages, emotion recognition in speech~\cite{asr-based-features-18-tits}, emotional or expressive speech synthesis~\cite{exploring_transfer_learning-19-tits, visualization-19-tits} or voice conversion~\cite{zhou2022emotional}.

Self-supervised learning is thus a specific form of Transfer Learning where a model is trained to learn representations of input data without the need for explicit supervision. 
These representations are the projection of the input data to a multidimensional space called latent space that captures information that is important for prediction of characteristics.

There is however still a lot work to do to understand how these latent spaces are structured, what characteristics can be predicted, how can they be disentangled, etc.

In this paper, we are particularly interested in providing a fine-grained expert annotations that can be aligned with a speech signal, allowing for exploration of relationships between speech representations and expert knowledge.

To this end, our rich phonetic annotations, augmented with syllable and stress information, serve as strong supervisory signals. Moreover, these phonetic transcriptions, tied to their written form, provide an explicit correspondence between the discrete symbols and their variably pronounced forms encountered in natural language. 
This could facilitate the discovery of speech units directly from the data. Hence, this research can provide valuable insights and push the boundaries of current methods in automatic speech recognition, synthesis, and analysis.

Conducting linguistic feature extraction, such as phonetic transcriptions, syllable separations, and word stress, plays an essential role in a multitude of fields, such as speech representation learning, speech synthesis~\cite{pradhan2013syllable, taylor1998architecture}, speech recognition, and speaker identification. The ability to accurately mark syllable boundaries in words is fundamental for understanding language structure and its phonetic variations, which in turn aids in efficient decoding and analysis of speech data. 

Among its potential use-cases, applications in the realms of second language learning and more specifically computer-assisted pronunciation training (CAPT) \cite{tits23_slate} can greatly benefit from the reliable extraction, ensuring the development of effective learning materials that enhance pronunciation and overall language proficiency in learners.

Nevertheless, the extraction of linguistic features poses challenges due to the inherent complexity and variability observed in natural languages. Dialectal variations, phonetic ambiguities, and inconsistencies in syllable boundaries are contributing factors that hinder the development of a reliable and consistent system for extracting linguistic features. Moreover, there is a lack of resources that offer consistent phonetic transcriptions encompassing stress marks, phone boundaries, and syllable boundaries across both pronunciation and spelling domains.

In this work, our goal is to define a methodology for linguistic feature extraction (phonetic transcriptions, stress marks, automatic syllabification in text and phonetic transcription domains) that is multilingual and compatible with forced-alignment tools. 
We have developed a process based on existing open-source building blocks that includes different steps and checks, as well as a consensus mechanism to extract the best possible linguistic features from text.

The Montreal Forced Aligner (MFA)~\cite{mfa-mcauliffe2017montreal} is an essential tool in our analysis for its function in phonetic alignment, providing detailed pronunciation transcriptions. It is important to note that, while MFA is commonly used to align audio signals with corresponding text transcriptions, we consider that task to be already efficiently handled by MFA's acoustic models. Our work aims to enrich this process: we focus on aligning phonetic syllabifications with graphemic representations of the corresponding words, essentially extracting and aligning units of sounds for precise syllabification across languages. We consciously designed our system to be fully compatible with the MFA, providing a complementary solution to the existing forced-alignment process.

By aligning phonetic syllabifications with their corresponding graphemic representations and creating a multimodal mapping, our methodology opens up new avenues of exploration in the field of speech representation learning. 

\begin{figure}[t]
  \centering
  \includegraphics[width=0.95\linewidth,height=0.85\linewidth]{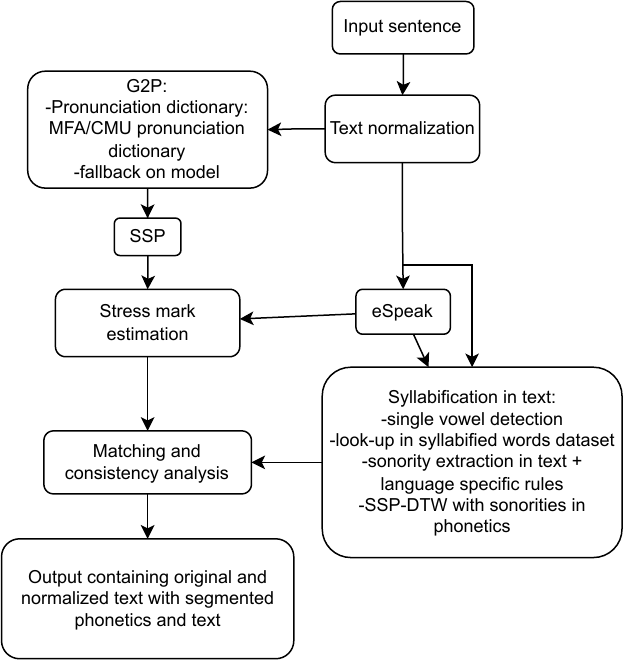}
  
  \captionsetup{skip=5pt,  belowskip=-15pt} 
  \caption{Block diagram of the linguistic feature extraction system described in Section~\ref{sec:system}}
  \label{fig:system}
\end{figure}

\section{Related Work}

Automatic syllabification is a challenging task for natural language processing due to the ambiguity of syllable boundaries. Different techniques have been developed to address this problem, including rule-based and data-driven approaches. In this section, we review some relevant studies on automatic syllabification in English, Spanish, Italian, and Portuguese.

For English, the study presented in~\cite{marchand2009automatic} compares five different algorithms, including two rule-based approaches and three data-driven techniques. The study finds that data-driven methods outperform rule-based systems in terms of word and juncture accuracy. Furthermore, syllabification in the pronunciation domain is easier than in the spelling domain. The study also highlights the challenge of establishing a gold standard corpus for evaluation due to the lack of consensus in the entries of multiple lexical databases. However, in their experiment, they apply the two rule-based algorithms in the spelling domain without any adaptation, and they do not consider the use of the Sonority Sequencing Principle.

The Sonority Sequencing Principle (SSP)~\cite{SSP-vennemann1987preference} is a widely used rule for syllabification, which states that syllables are formed by increasing then decreasing sonority. It is based on the sonority hierarchy, which assigns a relative sonority value to each phone. Vowels have the highest sonority, followed by approximants (such as /r/ and /w/), fricatives, nasals, and finally stops, which have the lowest sonority.
The linguistic literature identified exceptions to this principle, the main one being probably the sibilant-stop consonant cluster~\cite{iacoponi2011sylli, SSP-exceptions-yin2023frequent, SSP-exceptions-delisi2015sonority}. 
Implementations of the principle with processing of these exceptions been successfully applied for automatic syllabification in several languages in the pronunciation domain with very high word accuracies~\cite{bigi2010automatic, bigi2014generic, bigi2015automatic}. But it has also been applied in the spelling domain with some success for some languages.

In Spanish, \cite{hernandez2013automatic} points out that syllabification follows basic rules but may deviate due to various factors, such as diphthongs or hiatuses. Some variations in syllabification are also related to geographical and dialectal criteria. Therefore, automatic syllabification in Spanish requires taking into account these variations.
For Italian, \cite{iacoponi2011sylli} presents a rule-based method that uses the Sonority Sequencing Principle (SSP) and additional rules specific to Italian. The study evaluates their method on a dataset of sentences that were manually syllabified and reports an accuracy of 0.98-1 for some of the subjects. We could not find an application of SSP in the spelling domain in English. The reason is maybe because a naive application of SSP in the spelling domain would not perform very well.


Many data-driven syllabification methods using different levels of complexities of machine/deep learning models, that have the potential to be applied to several languages, have been developed but mainly for the phonetic domain only \cite{bartlett2009syllabification, rogova2013automatic, krantz2018syllabification, language-agnostic-neural_krantz2019language}.

In this literature review, we did not find any method that is capable of syllabification in both pronunciation and spelling domains and study the consistency between them.
In this work, we thus propose a methodology for a unified automatic syllabification and experiment it in several languages.




\section{System}
\label{sec:system}

The proposed methodology for linguistic feature extraction is illustrated in Figure~\ref{fig:system}. It includes several steps: text normalization, grapheme-to-phoneme (G2P) conversion, syllabification in the phonetic domain, and syllabification in the text domain. Lastly, a consistency analysis is conducted to identify words with inconsistent syllable counts, facilitating manual correction of the remaining exceptional cases.
The system is designed to be multilingual and compatible with forced-alignment tools, namely \textit{Montreal forced aligner} (MFA).

\subsection{Text normalization}

The initial stage of the process involves normalizing the text, which includes handling non-standard notations that differ from actual words. The system assumes that most punctuation symbols in English are attached to words, either at the end (commas, different kinds of dots, etc.) or at the start (double quotes can be at the start and end). For acronyms, the system assumes that they are written as a sequence of capital letters without dots between them. Numerals are translated to words using a rule-based algorithm with the Python library num2words\footnote{\url{https://pypi.org/project/num2words/}}.

\subsection{Grapheme-to-phoneme (g2p) conversion}

After normalizing the text, the system utilizes various methods to perform grapheme-to-phoneme (g2p) conversion. 
Phonetics is the study of the physical properties and production of speech sounds, while phonemics is concerned with the abstract and meaningful distinctions of sounds within a particular language, known as phonemes. Phonetics focus on the sounds themselves, while phonemics focus on the functional and linguistic aspects of those sounds. There exist different phonetic symbol sets categorizing speech sounds production (IPA, X-SAMPA, ARPAbet)

There is a language abuse in the state of the art of G2P models, as they are in fact performing the transformation of written language (graphemes) into a sequence of phonetic symbols (phones) and not phonemes. These terminologies are often used interchangeably in internet resources. In this paper we only work with phonetic transcriptions (sequence of phones).

First, it looks up the word in a pronunciation dictionary. If the word is not found, the system estimates its pronunciation using a machine learning model. This two-step methodology allows the system to use high-quality transcriptions from available dictionaries while handling the problem of out-of-vocabulary words with a machine learning model. However, this method is limited in that it cannot model dependencies of pronunciation on context. The system relies on manual human correction to handle this problem.

The system uses open-source resources as pronunciation dictionaries and fallback machine learning models, including the CMU pronunciation dictionary and an open-source CMU g2p model\footnote{\url{https://github.com/Kyubyong/g2p}}, as well as the MFA pronunciation dictionaries and their g2p models using a carefully described IPA phone set\footnote{\url{https://mfa-models.readthedocs.io/en/latest/mfa_phone_set.html}}.

\subsection{Syllabification in pronunciation domain (phonetic transcriptions)}

\label{sec:syllabification_pronunciation}
Syllabification in the phonetic domain is carried out by the system, employing the Sonority Sequencing Principle (SSP). The SSP is a well-accepted principle that states that syllables are formed by organizing sounds according to their sonority, which is a measure of the relative loudness or intensity of a sound. 

We based our implementation on SyllabiPy\footnote{\url{https://github.com/henchc/syllabipy}} github repository. We defined the sonority hierarchies for the different symbol sets used in this paper (CMU phone set\footnote{Based on the ARPABET phonetic symbol set: \url{https://en.wikipedia.org/wiki/ARPABET}}, MFA's IPA set, letters). Figure~\ref{fig:SSP_DTW} shows sonority curve examples for three words. The top curves are in the phonetic domain, while the bottom curves are in the spelling domain (see next section for explanations about the mapping between them).

The syllable breaks are determined by the local minima that have a vowel (sonority of value 5) located before themselves and after the last syllable break (or start of the word for the first syllable break). An additional rule is that a new syllable break cannot create a syllable that does not contain a vowel.

In the resources used as a basis, diphthongs are annotated as single phones, where hiatuses are annotate as two separate vowels. Therefore to correctly segment hiatuses, we represent all vowels by a sequence of two sonorities: 5, then 4. This allows us to generate a syllable breaks in case of hiatuses, without influencing the rest of the segmentation. In this case, the syllable breaks position will be placed after the vowel containing the local minimum. On the contrary the syllable breaks determined by consonant local minima will be placed before them.

The system handles sibilant-stop consonant clusters such as /skr/ and /spl/ thanks to the rule that a new syllable break cannot create a syllable that does not contain a vowel (mentioned earlier). 

As stress marks are not provided in MFA dictionaries and g2p models, we use eSpeak as an extra resource for retrieving this information. We compute a syllabified version of eSpeak transcription and extract stressed syllable index to augment the MFA transcription.

\subsection{Syllabification in spelling domain (text)}

 In the literature, it is commonly assumed that syllabification in the text domain results in a single, definitive number of syllables. However, pronunciation dictionaries, such as the CMU or MFA pronunciation dictionaries, provide variations of pronunciation, including variations in the number of vowels and, therefore, in the number of syllables.

To ensure consensus across datasets, we propose matching the number of syllables in text with the number of syllables in the pronunciation dictionary. This is consistent with the use of consensus as a valid mechanism for gathering data from manual annotators and was also used to combine datasets in~\cite{marchand2009automatic}.

We assume that the number of syllables is the same across variants of English. We proceed with syllabification in several steps. First, we detect if the word has only one vowel based on its phonetic transcription using the G2P section. This step increases accuracy and avoids imprecisions that may arise in the following steps.

The second step involves looking up the word in a publicly available corpus of manually syllabified words. For English, we use a dataset of manually syllabified words\footnote{\url{https://www.gutenberg.org/ebooks/3204}} from the Gutenberg Project. For French, we use the \textit{Lexique383}\footnote{\url{http://www.lexique.org/}} . We apply a systematic correction to group consonants alone in a word with the next syllable. This correction addresses the issue of the sC cluster mentioned in Section~\ref{sec:syllabification_pronunciation}.
For Spanish, we do not use any dataset and redirect everything to SSP.

\begin{figure}[t]
  \centering
  \includegraphics[width=0.9\linewidth,height=0.75\linewidth]{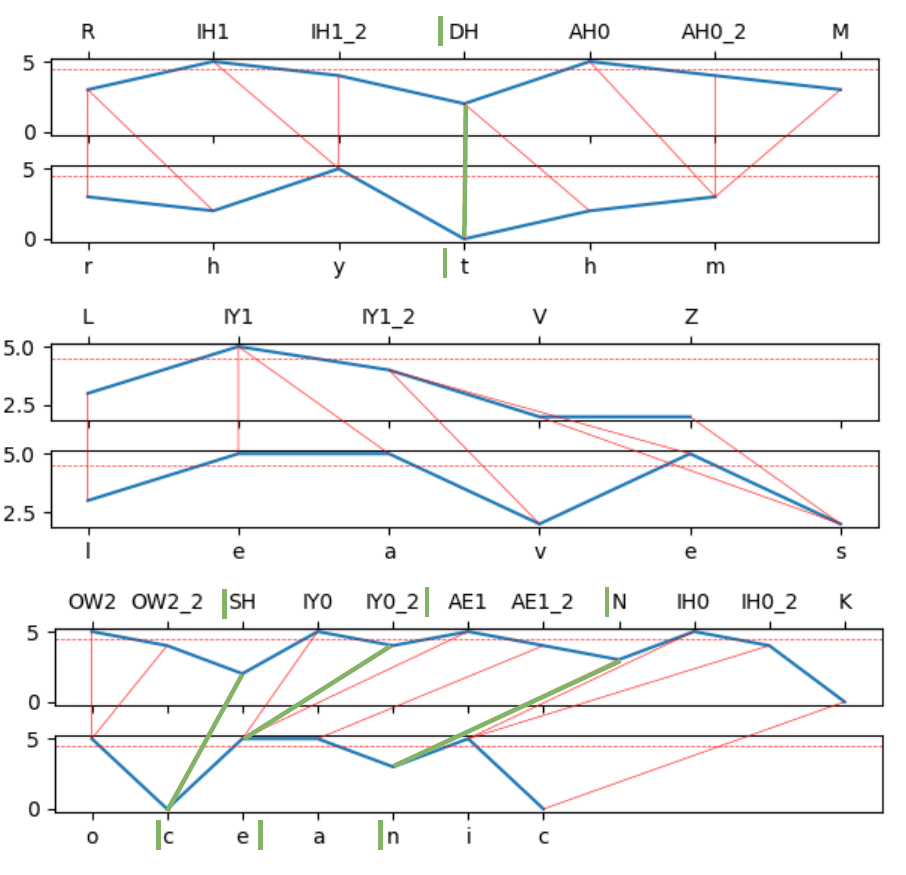}
  \captionsetup{skip=5pt,  belowskip=-15pt} 
  \caption{Illustration of the application of DTW on sonority sequences in the pronunciation and spelling domain. The blue curves are the sonority sequences, the red and green lines are the mapping links extracted from the DTW alignments. The green lines correspond to the local minima selected as syllable breaks in the phonetic domain and identifying the corresponding location in the spelling domain. The syllable break location are indicated with the vertical green pipe characters in both phonetic and spelling domains.}
  \label{fig:SSP_DTW}
\end{figure}

The third step involves processing words with more than one vowel that are \textit{out of vocabulary} (OOV). One could try applying SSP on the letters of the words, assuming the sonority of the letters. The performance of this method depends on the language. Specifically, this work well when the words follow a a predictable letter-to-sound mapping. To mitigate the limitation of this technique, it is also possible to add language specific rules.

However, SSP on text will struggle with hiatus, diphthongs, silent letters, and other cases for which the letter-to-sound mapping assumption is violated. To overcome this difficulty, we propose an approach that aligns sonority sequences in the pronunciation domain and the spelling domain using Dynamic Time Warping (DTW)~\cite{dtw-muller2007dynamic}. This approach allows us to benefit from the accurate prediction of syllable starts in the pronunciation domain and map them into the spelling domain. 

An illustration of this procedure is shown in Figure~\ref{fig:SSP_DTW} with three example English words containing cases where letter-to-sound mapping is not respected: (1) \textit{rhythm} contains a silent \textit{h}, a \textit{schwa} sound (symbol \textit{AH0} in CMU set) that does not correspond to a written letter, and a consonant sound written with two letters (\textit{th}); (2) \textit{leaves} containing the grapheme \textit{ea} as a single vowel, and a silent \textit{e}; (3) \textit{oceanic} containing the grapheme \textit{ea} as a hiatus.

\section{Experiments}


To evaluate the quality of an Automatic Syllabification algorithm, two measures are typically used: word accuracy and juncture accuracy. Word accuracy measures the proportion of words for which the number of syllables is exactly the same as a gold standard. Juncture accuracy measures the proportion of junctures that are the same as a gold standard.

In this study, we propose to measure word accuracy between the syllabified text of our methodology and the result of the application of the Sonority Sequencing Principle in the pronunciation domain. This is backed by the literature, as the number of syllables extracted in the phonetic domain is highly reliable. This measure allows for reproducibility and avoids comparison with a gold standard annotated by humans, which is also imperfect and inconsistent. 



Our consensus mechanism allowed us to detect errors that can complement syllabified text corpora or start corpora of edge cases for new languages.


\subsection{Distribution of number of syllables in words in natural language corpus and in a lexicon}

\begin{figure}[t]
  \centering
  \includegraphics[width=\linewidth,height=0.45\linewidth]{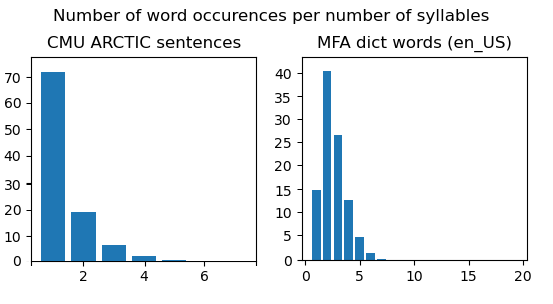}
  \captionsetup{skip=2pt,  belowskip=-15pt} 
  \caption{Proportions of words (in \%) in CMU ARCTIC sentences and MFA pronunciation dictionary per number of syllables in the word, according to sonority principle applied in the pronunciation domain}
  \label{fig:n_syl_hist}
\end{figure}

The word accuracy applied to sentences is not directly comparable to that of a lexicon of existing words in English. The reason for this is that the distribution of the number of syllables in a lexicon and in a set of sentences is very different. To illustrate this, Figure~\ref{fig:n_syl_hist} shows the proportion (in \%) of words for each possible number of syllables in CMU ARCTIC sentences and MFA pronunciation dictionary (en\_US variation).
The large proportion ($>70\%$) of single vowel words in sentences explains why the lexicon benchmarks are more challenging than a set of sentences. 

We therefore provide the results for both scenarios in Section~\ref{sec:ablation_words} and Section~\ref{sec:sentences}.

\subsection{Ablation Study on words}
\label{sec:ablation_words}

An essential step in our work involves the use of SSP for direct syllabification - a method we refer to as \textit{SSP}. It is pertinent to note that our implementation of this approach mirrors the implementation provided in the documentation of the Natural Language Toolkit (NLTK)\footnote{\url{https://www.nltk.org/api/nltk.tokenize.sonority_sequencing.html}}, a popular platform employed for multiple language processing tasks. NLTK's syllabification implementation also relies on SSP and supports various languages. This established baseline bears significance in our ablation study, where we gauge the additional contributions made by the other components of our methodology. The reader can directly spot the limitations of this method applied to text by consulting the given example in the link of the footnote with the word \textit{sentence}. Indeed, it is syllabified in 3 syllables (\textit{sen|ten|ce}), while it should be in 2 syllables (\textit{sen|tence}).

To measure the difference in performance between different languages, we performed an ablation study on English (variations GB and US based on MFA pronunciation dictionaries, as well as US with CMU pronunciation dictionary), Spanish, and French. We used a set of randomly selected 1000 words in the corresponding pronunciation dictionaries to report word accuracies in the different versions.

The first step of all the versions is the same and consists of single vowel checking through a look-up in the pronunciation dictionary.
Then, to be able to quantify the contributions of the technique of DTW between sonority sequencies of text and phonetics, and the contribution of using look-up in a dataset of syllabified words (when available), we compute word accuracies on 4 alternatives of the methodology, consisting in the possible component combinations:

\vspace{-7pt}
\begin{itemize}
  \setlength{\itemsep}{-2pt} 
\item \textit{SSP}: we directly use SSP on the letters, we use neither the DTW technique, neither look-up in the dictionary
\item \textit{lkp-SSP}: we first perform lookup in the syllabified words dataset to check if the word exist, and fallback on SSP on the letters
\item \textit{SSP-DTW}: extract sonority sequences and apply DTW to associate letters to phones and use SSP to extract starts of syllables
\item \textit{lkp-SSP-DTW}: we first perform lookup in the syllabified words dataset to check if the word exist, else we use SSP-DTW
\end{itemize}

\vspace{-10pt}



\begin{table}[h!]
\centering
\small 
\resizebox{\linewidth}{!}{
\begin{tabular}{|c|c|c|c|c|}
\hline
& SSP & lkp-SSP & SSP-DTW & lkp-SSP-DTW \\
\hline
es\_ES & 87.6 & - & \textbf{94.0} & - \\
fr\_FR & 82.3 & 85.9 & \textbf{90.1} & 89.1 \\
en\_GB & 88.5 & 94.4 & 92.6 & \textbf{95.5} \\
en\_US & 88.5 & 93.7 & 92.3 & \textbf{94.2} \\
CMU & 89.5 & 93.6 & 93.4 & \textbf{94.7} \\
\hline
\end{tabular}
}
\caption{Word accuracies for different language/variations and methods}
\label{table:ablation}
\end{table}

We report word accuracies for different versions of our methodology. The results are shown in Table~\ref{table:ablation}.
From the results, we can observe that the look-up in the syllabified words dataset has a positive effect over SSP (text only) for both French and English (all variations). We can also see that the SSP-DTW methodology performs better than the naive application of SSP on text, for all languages in our experiments.
For English, the highest accuracy is achieved by the \textit{lkp-SSP-DTW} version, indicating that the use of syllable corpus lookup in conjunction with DTW methodology can significantly improve the accuracy of automatic syllabification. This is however not true for experiments in French. This might indicate that the \textit{SSP-DTW} methodology is more reliable in itself than the human annotations collected in the dataset used for the experiment.


\subsection{CMU ARCTIC sentences}
\label{sec:sentences}

The CMU ARCTIC dataset \cite{cmu_arctic-kominek2004cmu} is a multi-speaker database consisting of 1132 phonetically balanced English utterances, recorded under studio conditions. The set of speakers include several accents of English.
The dataset was then generated by selecting a compact subset of utterances containing at least one occurrence of every diphone (phone pairs).

It was originally created to support speech synthesis research but it has been widely used in various applications since its release, including speech synthesis, voice conversion, speaker adaptation, prosody modeling, speech recognition, and linguistic studies. 
We therefore release the result of our unified phonetization and syllabification in text and phonetic domains to support future studies in these domains.
We also think that these annotations are useful information for speech representation learning as it could serve as data to analyze impact of contribution of different factors (speaker identity, accent, stress, rhythm), and potentially help in the disentanglement of these different factors.

Furthermore, other datasets including L2-ARCTIC~\cite{L2-ARCTIC-zhao2018l2}, and EmoV-DB~\cite{emov-dv-adigwe2018emotional} use the same transcriptions. L2-ARCTIC is a speech corpus of non-native English that is intended for research in voice conversion, accent conversion, and mispronunciation detection. 
The initial release of their dataset includes recordings from ten non-native speakers of English whose first languages are Hindi, Korean, Mandarin, Spanish, and Arabic, each L1 containing recordings from one male and one female speaker. Each speaker recorded
approximately one hour of read speech from the CMU ARCTIC sentences.
EmoV-DB consists of recordings of several speakers with different emotional categories in a parallel setup using CMU ARCTIC sentences. These sentences do not convey particular emotions in the text which would help to disentangle emotional expressiveness in speech from the textual content.

The phonetization and unified syllabification described in Section \ref{sec:system} was applied to the 1132 CMU ARCTIC sentences. The word accuracy obtained on all the words is $>99.8\%$.

\section{Conclusions}

This study introduced a novel, multilingual methodology for linguistic feature extraction, designed to be compatible with forced-alignment tools. Our approach effectively extracted essential linguistic features, including phonetic transcriptions, stress marks, and automatic syllabification in both text and phonetic domains. The methodology integrated various techniques, such as text normalization, grapheme-to-phoneme conversion, syllabification in the phonetic and text domains, and a consensus analysis to identify inconsistencies.

Our ablation study demonstrated the efficacy of the proposed methodology in automatically syllabifying words across multiple languages. The optimal performance was achieved by combining corpus lookup and Dynamic Time Warping (DTW) on sonority sequences. This approach can be further enhanced by progressively incorporating edge cases into the training dataset.

By applying our methodology to the CMU ARCTIC dataset, we generated valuable data that can benefit various speech-related research domains, available online\footnote{\url{https://github.com/noetits/MUST_P-SRL}}. Our unified phonetization and syllabification annotations have the potential to advance speech representation learning and disentangle different factors in speech technologies, such as speech synthesis and speech analysis tasks.

\section*{Limitations}


This paper concentrates on the intersection of phonetics and syllabification, aiming to align phonetic transcriptions with corresponding graphemes. While we mention the term \textit{alignment}, the context in this paper refers to the alignment of phonetic transcriptions with their corresponding graphemes, a pivotal step in our methodology for accurate multilingual syllabification. Highlighting this nuance provides a correct understanding of the terminologies and approaches used in this study, and sheds light on the specific challenges and contributions of our work.

Future research directions include extending the proposed methodology to additional languages and investigating the impact of our linguistic feature extraction on specific speech technology applications. Furthermore, refining the methodology by incorporating language-specific rules or addressing limitations in the consensus analysis could lead to even more accurate and robust results.

While our methodology presents improvements in linguistic feature extraction and automatic syllabification, some limitations should be noted. Firstly, while we aimed to create a multilingual system, our current implementation and evaluations were focused mainly on English, French, and Spanish. Extending and evaluating our methodology across other languages, especially those with vastly different phonetic structures, remains a future challenge. 

Secondly, the system heavily relies on the availability and quality of pronunciation dictionaries for its grapheme-to-phoneme conversion process. As such, issues like handling out-of-vocabulary words or modeling pronunciation dependencies based on context heavily depend on manual correction, limiting the scalability of the system. Note however that the choice of MFA tools was done among other things because of the large list of languages it supports (see the pronunciation dictionaries\footnote{\url{https://mfa-models.readthedocs.io/en/latest/dictionary/index.html}} and g2p models\footnote{\url{https://mfa-models.readthedocs.io/en/latest/g2p/index.html}}).

Thirdly, our approach to identifying and addressing inconsistencies between different syllabification resources uses a consensus mechanism which, while effective, may still retain inaccuracies inherent in these resources.

Acknowledging these limitations provides valuable directions for potential future enhancements and research towards fully automated and accurate linguistic feature extraction.


\section*{Acknowledgements}

This work is part of the project \textit{REDCALL} that is partially funded by a FIRST Entreprise Docteur program from SPW Recherche\footnote{https://recherche.wallonie.be/}

This project is a collaboration between Flowchase SRL and the Information, Signal and Artificial Intelligence Lab (ISIA Lab) of University of Mons in Belgium.

\bibliography{anthology,custom}
\bibliographystyle{acl_natbib}

\appendix



\end{document}